# Detection of a facemask in real-time using deep learning methods: Prevention of Covid 19

**Gautam Siddharth Kashyap[1*], Jatin Sohlot[2], Ayesha Siddiqui[3], Ramsha Siddiqui[4], Karan Malik[5], Samar Wazir[6], Alexander E. I. Brownlee[7]**

[1]officialgautamgsk.gsk@gmail.com, [2]j16sohlot@gmail.com, [3]ayeshamasood2912@gmail.com, [4]rmasood0129@gmail.com, [5]karanmalik2000@gmail.com, [6]samar.wazir786@gmail.com, [7]sbr@cs.stir.ac.uk

[1,6]Department of Computer Science and Engineering, SEST, Jamia Hamdard, New Delhi, India
[2]School of Computer Science and Engineering, VIT University, Vellore, India
[3,4]Friedrich-Alexander-Universität Erlangen-Nürnberg, Germany
[5]Arizona State University, Tempe, Arizona, USA
[7]Division of Computing Science & Mathematics, University of Stirling, UK
*Corresponding Author: Gautam Siddharth Kashyap

*Abstract*— **A health crisis is raging all over the world with the rapid transmission of the novel-coronavirus disease (Covid-19). Out of the guidelines issued by the World Health Organisation (WHO) to protect us against Covid-19, wearing a facemask is the most effective. Many countries have necessitated the wearing of face masks, but monitoring a large number of people to ensure that they are wearing masks in a crowded place is a challenging task in itself. The novel-coronavirus disease (Covid-19) has already affected our day-to-day life as well as world trade movements. By the end of April 2021, the world has recorded 144,358,956 confirmed cases of novel-coronavirus disease (Covid-19) including 3,066,113 deaths according to the world health organization (WHO). These increasing numbers motivate automated techniques for the detection of a facemask in real-time scenarios for the prevention of Covid-19. We propose a technique using deep learning that works for single and multiple people in a frame recorded via webcam in still or in motion. We have also experimented with our approach in night light. The accuracy of our model is good compared to the other approaches in the literature; ranging from 74% for multiple people in a nightlight to 99% for a single person in daylight.**

## I. INTRODUCTION

In the incipient phases of Covid-19, the world health organization (WHO) advised countries to make sure that their citizens wear a mask in public places, due to the large number of positive cases arising from crowded places. When the number of infected people increased, the WHO declared it a global pandemic. The mask was used by only a few people before this pandemic to protect themselves from air pollution. The potential of the mask is to reduce the risk and vulnerability from the infectious individuals during the pre-symptomatic period. Therefore, the detection of facemask plays an important role in this global pandemic. For the prevention of Covid-19, artificial intelligence techniques like machine learning and deep learning methods can be used [1]. These techniques can be used to design a helpful system to predict and monitor the spread of the virus. For example, in France, AI software is implemented in surveillance cameras of the Paris metro to monitor whether the people are wearing a mask or not [2]. To tackle Covid-19, many researchers have gained interest and contributed recently by giving various monitoring, screening, and compliance-assessment applications [3]. Software developed by French startup Dataka Lab detects those people who are not wearing a mask. The main aim of this software is to collect the statistical data of the people who are not wearing a mask and help the authorities by predicting the potential outbreak of Covid-19 [4].

Facemask detection contains two types of problems:

1. Detection problem: In digital images, it detects the location of the faces of people.
2. Classification problem: Detects whether the facemask is present or not on the detected faces.

There are many studies related to the detection problem in the computer vision literature [5]. On the other hand, the classification problem has recently gained interest when the Covid-19 pandemic hit the world. The increasing number of Covid-19 cases inspired the computer vision research community to develop methods to detect facemask in a public place and to contribute to constraining the Covid-19. Many studies on facemask detection in public places were published last year [6]–[23]. However, many of these studies do not test their proposed approaches on either multiple people in a single frame or relatively darker lighting conditions (nightlight). Moreover, real-time testing of the systems has also been overlooked in some cases.

The dataset we used to train our Convolutional Neural Network (CNN) is made by combining two existing datasets - the Masked FaceNet [24] and the Face Mask Dataset [25]. Our dataset, the Masked Face Detection trio (MFD_trio) contains 3 categories, namely correct_mask, incorrect_mask, and without_mask. The dataset has been explained in detail in section III. For real-time implementation, we have used webcam images as well as videos to test our model for single and multiple people in a frame, for both daylight and nightlight.

The primary objective of our proposed technique is as follows:

1. Accurately classify faces within images as having no/incorrect/correct wearing of a facemask.
2. Good performance under varied and challenging conditions (low lighting, multiple faces, etc.)
3. Fast run time (<10ms)
4. Promoting the use of facemask in public places with the help of deep learning methods.
5. With the help of a webcam, we have tried to create a real-time scenario so that our proposed technique can work well.
6. Ensure the safety of the working environment.

The paper is structured as follows. The next section i.e. Section II consists of the literature review on existing datasets and facemask detection techniques. In section III the proposed datasets have been explained. In section IV implementation of our approach has been presented. In section V the result of the experimentation is discussed. In section VI we conclude our paper with possible improvements and future scope.

## II. LITERATURE REVIEW

Many studies have been done in developing and improving deep learning methods, mainly in object detection and recognition for varying application domains [26], [27]. Ge et al. [28] suggests that all intents and purposes of face detection have the following two challenges:-

1. Unavailability of large datasets containing both the masked and unmasked faces.
2. Omission of facial expression in the covered region.

In the last year, quite some appreciable work has been done on the problem of facemask detection. In this section, we will review some research related to our proposed work to provide a background. Firstly, we will discuss the existing datasets and then we will discuss the existing techniques for facemask detection.

### A. Facemask Datasets

The problem of masked face detection demanded little attention from the Computer Vision community before the Covid-19. Rather masked faces were generally studied within a wider problem domain of processing and classifying close-up facial images [29]. Based on popularity and public availability for research, we identified the following datasets:-

Wang et al. [3] improved the performance of face recognition methods by introducing the following datasets:-

MFDD (Masked Face Detection Dataset):-

1. It is an extended version of the dataset which was initially released by [30]. This dataset contains 24,771 masked faces.
2. This dataset contains face images taken in challenging conditions from all over the internet. Its primary use is for training purposes and evaluating the facemask detectors.

RMFRD (Real World Masked Face Recognition Dataset):-

1. This dataset contains 5000 masked faces and 90000 normal faces of 525 people.
2. This dataset has a drawback as some of the images of people labelled as wearing masks are with incorrectly worn masks. Hence it is not recommended to use this in a global pandemic.

SMFRD (Simulated Masked Face Recognition Dataset):-

1. This dataset is a simulated dataset.
2. In this dataset, artificial masks are added to the faces of images. It contains 500,000 masked faces.

Ge et al. [28] proposed the MAFA dataset (i.e. Masked Faces) but the term masked used here generally does not include only properly masked face images, but also includes images in which the face is blocked or covered by some other object. It collected images from all over the internet. This dataset contains 30,811 images.

The above-discussed dataset does not identify masked faces with a correctly and incorrectly worn mask. Two datasets can be found on the Kaggle portal which makes the difference [31], [32]. These are:-

FMD (Face Mask Detection):-

1. This dataset contains 853 images which are divided into three categories that are with mask, without a mask, and mask worn incorrectly.
2. There are 4072 annotated images out of which 3232 are mask images, 717 are without mask images, and 123 are incorrectly worn mask images.
3. The limitation of these datasets is that they are too small which limits the use for the improvement and evaluation of facemask detectors.

MMD (Medical Mask Dataset):-

1. This dataset contains 6024 images collected from unconstrained settings.
2. In this dataset, 20 classes of images are labelled as correctly masking, incorrectly mask, no mask, etc, and 16 classes are reserved for masks, face shields, hats, etc.

3. This dataset has 9067 annotated faces out of which 6758 are mask images, 2085 are without mask images, and 224 are incorrectly worn mask images.

Batagelj et al. [33] developed a dataset by combining two datasets MAFA and Wider face and called it FMLD (Face Mask Label Dataset). This approach aims to provide already defined data with accompanying labels to train, test, and compare the facemask detection models to detect if the mask is present or not. A summary of all the above datasets is provided in Table I.

TABLE I
DATASETS SUMMARY

| Dataset | Dataset Type | Mask Type | Number of Images | Number of Faces | Available Labels | | |
|---|---|---|---|---|---|---|---|
| | | | | | Mask | No mask | Inc. worn |
| MFDD | Image | Real-world | 4343 | N/A | 24771 | 0 | 0 |
| RMFRD | Image | Real-world | 92671 | 92671 | 2203 | 90000 | 0 |
| SMFRD | Image | Stimulated-world | 500000 | 500000 | 500000 | 0 | 0 |
| MAFA | Image | Real-world | 30811 | 36797 | 35806 | 991 | 0 |
| FMD | Image | Real-world | 853 | 4072 | 3232 | 717 | 123 |
| MMD | Image | Real-world | 6024 | 9067 | 6758 | 2085 | 224 |
| FMLD | Annotated | Real-world | 41934 | 63072 | 29532 | 32012 | 1528 |

Note that the number of reported images is different from the actual numbers publicly available online.

## B. Facemask Detection Techniques

The following works on facemask detection techniques were studied:-

Ejaz et al. [6] solved the masked and unmasked facial recognition using principal component analysis(PCA). Their proposed technique, to identify the face without masks has 96.25% accuracy on the other hand while detecting the face with a mask, their accuracy gets decreased to 68.75% accuracy. Li et al. [7] proposed the YOLOv3 for the detection of the face. YOLOv3 is based on a darknet-19 architecture which is a deep learning network. The authors used the Wider face and CelebA datasets for training, and FDDB datasets were used for evaluation purposes. Their model achieved an accuracy of 93.9%. Das et al. [8] used simple machine learning packages like TensorFlow, Keras, OpenCV, etc. to detect whether the facemask is present or not on the image. They have used two datasets and they achieved an accuracy of 95.77% and 94.58% respectively. Loey et al. [9] used the ResNet 50 for feature extraction and used the SVM, ensemble algorithms, and decision trees to detect facemask. They have used the three datasets RMFRD, SMFD, and LFW and achieved a testing accuracy of 99.64%, 99.49%, and 100% respectively. Inamdar and Mehendale [10] developed a facemask detection system using the deep learning methods for the three categories, which are facemask correctly worn, incorrectly worn, and no facemask at all. The authors claimed that they achieved an accuracy of 98.6%. Qin and Li [11] used the SRCNet classification network for the detection of the facemask. The authors claimed that their proposed technique achieved an accuracy of 98.7%. Li et al. [12] developed a system to detect whether the facemask is present or not by using the HGL method for head pose classifications with facemasks. The authors claimed that they achieved an accuracy of 93.64% from the front and a side accuracy of 87.17%. Khandelwal et al. [13] developed a system to detect whether the facemask is present or not by using the deep learning model and converting the image into black and white. The model achieved a 97.6% accuracy. There are a few limitations to this model that if the camera height is above 10 feet then it will not be able to detect faces as well as it could not detect partially hidden faces. Jiang et al. [14] proposed a retina facemask. They have used the ResNet and MobileNet models on large datasets with 7959 images with the help of transfer learning to extract robust characteristics. Ristea et al. [15] developed a system to detect facemasks from a speech by using the multiple ResNet models and have trained GAN to do binary classification. Ud Din et al. [16] proposed a technique to reconstruct the area covered by the facemask and for the removal of the facemask using GAN-based network architecture. Rodriguez et al. [17] proposed a model for the automatic detection and identification of mandatory surgical facemasks in operating rooms. The main aim of this model is to trigger the alarm when the staff is not wearing the facemask. The authors claimed that their model achieved an accuracy of 95%. Matthias et al. [18] developed a facemask recognition project to detect if the facemask is present or not by capturing the images in real-time. They have used decision-making algorithms. Baidu et al. [19] developed an open-source tool to detect faces without facemasks. Their model is based on Pyramid Box that was published by the author in 2018 [20]. Li et al. [21] used the CNN model for the speedy face detection that evaluates low resolution an input image and eliminates nonface sections because of this high precise detection is possible. The authors have used the calibration nets to simulate detection. This model is fast as it achieves 14 FPS on CPU and can be increased to 100 FPS on GPU. Farfade et al. [22] developed a face detection system to solve the problem of the multiview face detection system. The proposed model is simple and can recognize faces at numerous angles. Jignesh Chowdary et al. [23] used the deep learning model InceptionV3 to detect whether the facemask is present or not. The authors claimed that their model achieved an accuracy of 99.9% during testing and 100% on training on the SMFD dataset. A limited number of studies had been reported on facemask detection. We had tried to cover it all, but further improvement is still required to the existing methods. This paper makes a step towards filling these gaps such as multiple face detection, dark environment testing, or real-time operation. In contribution to the fight against novel-coronavirus disease(Covid-19), we propose a technique for the detection of a facemask in real-time using deep learning methods. In the next section, we will see the types of datasets that we will use. A summary of all the above facemask detection techniques is provided in Table II.

TABLE II
FACEMASK DETECTION TECHNIQUES

| Authors | Proposed Methodology Used | Accuracy Achieved | Datasets Used | Limitations |
|---|---|---|---|---|
| Ejaz et al. [6] | PCA | Without mask-96.25% <br><br> With mask-68.75% | ORL Face Dataset [34] and some additional images are also added to the datasets | Their proposed work is better for the normal face but not for masked face recognition. |
| Li et al. [7] | YOLOv3 | 93.9% | Wider face [35], CelebA [36], and FDDB datasets [37] | The authors had tested their proposed approach on synthetic datasets only, real-time evaluation of the proposed model is not done. Subsequently, larger face datasets and other models can also be used to check the performance. |
| Das et al. [8] | TensorFlow, OpenCV, and Keras | 95.77% and 94.58%(on two datasets respectively) | FMD dataset [31] | Their proposed model fails to detect faces when they are in motion. Also, there is a limitation of these datasets is that they are too small which limits the use for the improvement and evaluation of facemask detectors. |
| Loey et al. [9] | ResNet 50 with some classical machine learning approaches | Testing Accuracy(RMFD)-99.64% <br><br> Testing Accuracy(SMFD)-99.49% <br><br> Testing Accuracy(LFW)-100% | RMFRD [3], SMFD [3], and LFW [38] dataset | RMFRD dataset has a drawback as some of the images of people labelled as wearing masks are with incorrectly worn masks. Hence it is not recommended to use this in a global pandemic. Therefore results can fluctuate if we remove the RMFRD dataset from their proposed model. |
| Inamdar and Mehendale [10] | Facemasknet | 98.6% | Authors had created their datasets of 35 images i.e. 10 for correctly mask, 15 for incorrectly worn, and 10 for no mask at all. | The authors had tested their proposed approach on very fewer size datasets because of that the chance of error gets decreased. Therefore to evaluate the real performance of the model a large size dataset is needed. |
| Qin and Li [11] | SRCNet | 98.70% | MMD dataset [32] | Their proposed works take a long time in identifying a single image. Hence it is not suitable for a video stream. |
| Li et al. [12] | HGL method | Front accuracy-93.64% and Side accuracy-87.17% | MAFA dataset [28] | The authors had not tested their proposed work for incorrectly mask worn conditions. |
| Khandelwal et al. [13] | MobileNetV2 | 97.6% | Authors had created their datasets of 4225 images i.e. 1900 for the mask, and 2300 for no mask at all, and 25 are of low-quality images. | There are a few limitations to this model that if the camera height is above 10 feet then it will not be able to detect faces as well as it could not detect partially hidden faces. |
| Jiang et al. [14] | ResNet and MobileNet | --------- | Wider face [35], FMD [31], and MAFA dataset [28] | In their proposed work there is a limitation on the performance of the dataset because other components do not work well with ResNet only Imagenet can be seen working. |
| Ristea et al. [15] | ResNet | 74.6% | MSC [39] | The authors had not tested their proposed work for incorrectly mask worn conditions also the accuracy of the proposed model is not good. |
| Ud Din et al. [16] | GAN-based Network | --------- | The synthetic dataset created using CelebA [36] | Their proposed works fail to detect the mask objects when the shape, colour, and size of the mask objects are different from the synthetic dataset. |
| Rodriguez et al. [17] | Face and Mask Detectors Combined | 95% | BAO datasets [38] and some additional images are also added to the datasets | The authors had not tested their proposed work for incorrectly mask worn conditions. |
| Matthias et al. [18] | decision-making algorithms | --------- | RMFRD dataset [3] | This dataset had a drawback as some of the images of people labelled as wearing masks are with incorrectly worn masks. Hence it is not recommended to use this in a global pandemic. |

Table II(continued...)

| Authors | Proposed Methodology Used | Accuracy Achieved | Datasets Used | Limitations |
|---------|---------------------------|-------------------|---------------|-------------|
| Baidu et al. [19] | Artificial Intelligence | 96.5% | The software creates its datasets. | This software needs a high-quality camera for correct recognition of the face as well as it had some optimization issues. |
| Li et al. [21] | CNN | --------- | FDDB [37] and AFW dataset [40] | Their proposed work well on AFW dataset as it is less in size but its performance gets decreased on the FDDB dataset. |
| Farfade et al. [22] | Deep Dense Face Detector | --------- | AFLW [41], FDDB [37], and PASCAL dataset [42] | Their proposed work cannot detect the faces when there are occluded and rotated faces. |
| Jignesh Chowdary et al. [23] | InceptionV3 | Training Accuracy(SMFD)- 99.9% Testing Accuracy(SMFD)- 100% | SMFD dataset [3] | The authors had not tested their proposed work for incorrectly mask worn conditions. |

## III. DATASETS

We have combined 2 datasets for training our model. These are:

1. Masked FaceNet [24] - This dataset contains 67,049 images of people wearing a facemask correctly and 66,734 images of incorrect masks i.e. nose, mouth, or chin uncovered. Due to computational limitations, we have used 950 and 928 images from each category for training, and 100 and 96 images each for testing.

2. Face Mask Data [25]- We have used this dataset for the third category i.e. no mask. This dataset contains 656 images of people without a mask for training and 97 in the test set.

In total, we have 2534 images in the training set for the 3 categories and 293 images for the test set.

For real-time application, we test our model in different lighting conditions - daylight and nightlight with single and multiple people in each frame. The following were the test categories for the real-time implementation:

- In Daylight
1. WebCam_Images_Single_Day(i.e. images clicked via webcam, containing a single person in daylight).
2. WebCam-Images_Multiple_Day(i.e. images clicked via webcam, containing multiple people in daylight).
3. WebCam_Videos_Single_Day(i.e. video recorded via webcam, containing a single person in daylight).
4. WebCam-Videos_Multiple_Day(i.e. video recorded via webcam, containing multiple people in daylight).

- In Nightlight
1. WebCam_Images_Single_Night(i.e. images clicked via webcam, containing a single person in nightlight).

2. WebCam-Images_Multiple_Night(i.e. images clicked via webcam, containing multiple people in nightlight).
3. WebCam_Videos_Single_Night(i.e. video recorded via webcam, containing a single person in nightlight).
4. WebCam-Videos_Multiple_Night(i.e. video recorded via webcam, containing multiple people in nightlight).

## IV. IMPLEMENTATION

Our proposed implementation uses a Convolutional Neural Network(CNN) to classify amongst 3 categories:

1. People wearing masks properly
2. People wearing masks incorrectly i.e. nose and/or mouth exposed
3. People not wearing a mask

In our dataset, we have 950, 928, and 656 training images for each of our 3 categories respectively, while in the testing phase, the number of images is 100, 96, and 97 respectively for each category.

Our self-created CNN takes images of size 150*150*3 pixels as input. This input layer is followed by 3 pairs of Convolutional and Max Pooling layers. Each of the 3 Convolutional layers has 32 filters with a kernel size of 3*3. Each layer uses "valid" padding and a stride of(1,1). The Rectified Linear Unit(relu) activation function is applied after each Convolutional layer.

The output from the final Max Pooling layer is sent to the Flatten layer, which flattens its input of size(n, 17, 17, 32) to a vector with shape (n, 17*17*32) i.e.(n, 9248). This vector now acts as an input to the 2 Dense layers. The first Dense layer has 100 nodes with the relu activation. The final layer is a Dense layer with 3 nodes, and the Softmax activation function. Each of the 3 nodes represents a classification category - correct_mask, incorrect_mask, and without_mask. The final model has 944,595 parameters, and all of these parameters are trainable. Figure (1) shows the structure and summary of the model used.

```
Model: "sequential_1"

Layer (type)                 Output Shape              Param #
=================================================================
conv2d_1 (Conv2D)            (None, 148, 148, 32)      896

max_pooling2d_1 (MaxPooling2 (None, 74, 74, 32)        0

conv2d_2 (Conv2D)            (None, 72, 72, 32)        9248

max_pooling2d_2 (MaxPooling2 (None, 36, 36, 32)        0

conv2d_3 (Conv2D)            (None, 34, 34, 32)        9248

max_pooling2d_3 (MaxPooling2 (None, 17, 17, 32)        0

flatten_1 (Flatten)          (None, 9248)              0

dense_1 (Dense)              (None, 100)               924900

dense_2 (Dense)              (None, 3)                 303
=================================================================
Total params: 944,595
Trainable params: 944,595
Non-trainable params: 0
```

Fig. 1. Summary of the CNN used to classify among the 3 categories

The input images are preprocessed and augmented as follows:

1. They are rescaled into the range [0,1].
2. Shearing transformation is applied to the images.
3. Random parts of the image are zoomed in to make the model more resilient.
4. Finally, each of the images is flipped horizontally.

The model is trained for 10 epochs with a batch size of 16 images. This trained model is saved as an h5 file and used for making predictions and testing.

OpenCV is used for accomplishing our task of real-time detection of facemasks. The image or video stream is sent as input to the system. The individual frames from the video are extracted and all the faces from each of the frames are detected using the pre-trained Haar Cascade Frontal Face Object Detection model. These faces are then sequentially extracted and preprocessed so that they can be used as input to our saved model. These preprocessed facial images are then inputted to the saved CNN and it classifies them into 1 of the three predefined categories.

A green rectangle appears surrounding the faces belonging to category 1 i.e. the mask is correctly worn, a blue rectangle appears on faces belonging to category 2 i.e. mask incorrectly worn and a red rectangle is drawn surrounding those belonging to the third category i.e. no mask. Since the model uses facial detection before the prediction process, it can detect facemasks for multiple people in a single frame.

As described in Section III, the model has tested for each of the 4 cases in both daylights as well as a nightlight.

## V. RESULT

Using the 2534 images belonging to the 3 categories for training our model, we achieved a training accuracy of **98.9%**. On the 293 images in the test set, the model managed to get an accuracy of **98.74%**.

Further, the real-time detection capabilities of the created system were tested using a live webcam stream, in both daylight as well as a nightlight, for 4 cases each:

1. Webcam images with a single person
2. Webcam images with multiple people
3. Webcam video with a single person
4. Webcam video with multiple people

The model performed well for all the cases in both lighting conditions. The classification in all the cases is completed within milliseconds, thus there is no apparent lag between the actual video and the prediction. Figures 2(a) to 2(d) and 3(a) to 3(d) show glimpses of the model testing in daylight and nightlight respectively:

- Figures 2(a) and 3(a): Webcam images of a single person in daylight and nightlight respectively.
- Figures 2(b) and 3(b): Webcam images of multiple people in daylight and nightlight respectively.
- Figures 2(c) and 3(c): Frame from webcam video of a single person in daylight and nightlight respectively.
- Figures 2(d) and 3(d): Frame from webcam video of multiple people in daylight and nightlight respectively

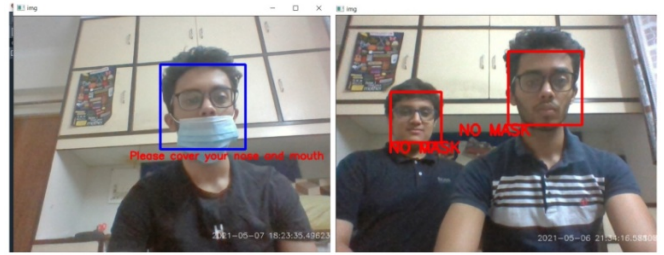

Fig. 2a & 2b. Webcam images of a single person and multiple people in daylight

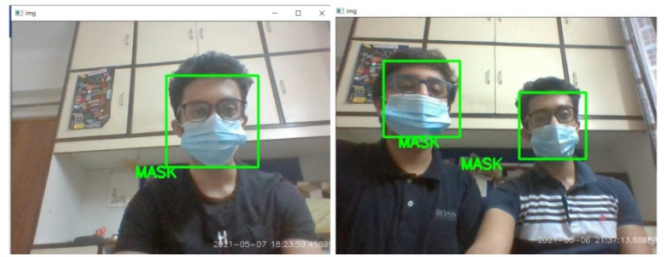

Fig. 2c & 2d. Frame from webcam video of a single person and multiple people in daylight

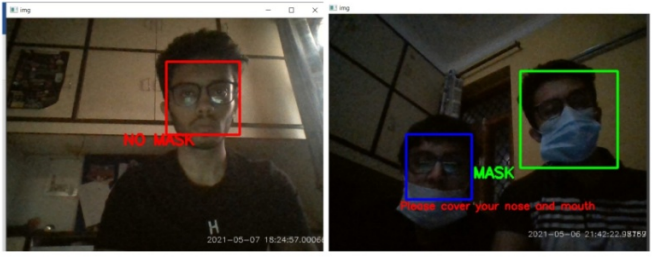

Fig. 3a & 3b. Webcam images of a single person and multiple people in nightlight

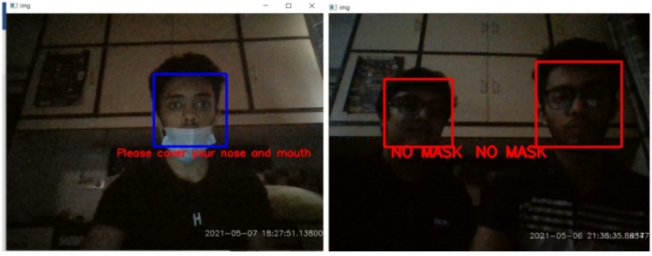

Fig. 3c & 3d. Frame from webcam video of a single person and multiple people in nightlight

Table III shows the accuracy of the model in different lighting conditions when it is tested with 50 images each having single and multiple people. These images are available can be accessed using the link available in Section VII of our paper. Figure (4) shows the classification report of the proposed model.

TABLE III
MODEL ACCURACY FOR 50 IMAGES EACH IN DAYLIGHT AND NIGHTLIGHT WITH SINGLE AND MULTIPLE PEOPLE

| Lighting Condition | Accuracy with a single person | Accuracy with multiple people |
|---|---|---|
| Daylight | 99% | 98% |
| Nightlight | 88% | 74% |

```
                precision    recall  f1-score   support

  correct_mask       0.99      0.99      0.99       100
incorrect_mask       0.98      0.99      0.98        96
 without_mask        1.00      0.99      0.99        97

      accuracy                          0.99       293
     macro avg       0.99      0.99      0.99       293
  weighted avg       0.99      0.99      0.99       293
```

Fig. 4. Classification report of the proposed model

## VI. CONCLUSION & FUTURE SCOPE

In this paper, we proposed a technique for detecting facemasks in a real-time scenario using deep learning methods. We have combined two datasets, as specified in the Dataset section. A Convolutional Neural Network (CNN) was trained on 3 categories of images - namely correct_mask, incorrect_mask, and without_mask. With 10 epochs, we achieved a training accuracy of 98.9%. The testing accuracy came out to be 98.74% for the 293 images in our test set.

The model was then integrated with Haar Cascade facial detection using OpenCV for detecting facemasks in real-time. The integrated system was tested with images and videos having single and multiple people in both daylights as well as a nightlight. The accuracy in these cases was approximate:

1. 99% for a single person in a frame in daylight
2. 98% for multiple people in a frame in daylight
3. 88% for a single person in a frame in nightlight
4. 74% for multiple people in a frame in nightlight

Our model is competitive with the other published techniques in terms of accuracy. Moreover, our proposed technique is less complex in structure and in working and gives faster results. It can be easily integrated with CCTV cameras or in drones to monitor a large number of people for identification of facemask and prevention of Covid-19 and other airborne diseases.

However, the accuracy of our model can be improved in nightlight, possibly by using better quality cameras. Further, it can be improved in identifying the types of masks that a person is using and also to identify the colour of the mask in the real-time scenario.

## VII. DECLARATIONS

### A. Funding: Not Applicable

### B. Conflict of Interest: Not Applicable